\documentclass[runningheads]{llncs}
\usepackage[T1]{fontenc}
\usepackage{graphicx,verbatim}
\usepackage{times}
\usepackage{soul}
\usepackage{url}
\usepackage[hidelinks]{hyperref}
\usepackage[utf8]{inputenc}
\usepackage[small]{caption}
\usepackage{amsmath}
\usepackage{svg}

\usepackage{amsthm}
\usepackage{amssymb}
\usepackage{booktabs}
\usepackage{algorithm}
\usepackage{algorithmic}
\usepackage[switch]{lineno}
\usepackage{multirow}
\usepackage{xcolor}
\usepackage{colortbl}
\usepackage{array}

\begin{document}
\title{Motion-Guided Causal Disentanglement for Robust Multi-View Cine Cardiac MRI Diagnosis}

\author{%
Chuankai Xu\inst{1} \and
Cristiane De Carvalho Singulane\inst{1} \and
Mohammad Abuannadi\inst{1} \and
Stephen Chandler\inst{1} \and
Jeremy Slivnick\inst{2} \and
Karolina Zareba\inst{3} \and
Jane Cao\inst{4} \and
Vidya Nadig\inst{5} \and
Fabio Fernandes\inst{6} \and
Seth Uretsky\inst{7} \and
Diego Perez de Arenaza\inst{8} \and
Amit Patel\inst{1} \and
Jianxin Xie\inst{1}}
\authorrunning{C. Xu et al.}
\institute{%
University of Virginia, Charlottesville, VA, USA\\
\email{\{rzv4ve,hcf7fd\}@virginia.edu} \and
University of Chicago, Chicago, IL, USA \and
Ohio State University, Columbus, OH, USA \and
St.\ Francis Hospital \& Heart Center, Roslyn, NY, USA \and
Hartford HealthCare, Hartford, CT, USA \and
Instituto do Cora\c{c}\~ao (InCor), S\~ao Paulo, Brazil \and
Atlantic Health System, Morristown, NJ, USA \and
Sociedad Italiana de Beneficencia (Hospital Italiano), Buenos Aires, Argentina}
  
\maketitle
\begin{abstract}
Multi-view cardiac magnetic resonance (CMR) imaging provides complementary anatomical information and is widely used for noninvasive disease assessment. Recent transformer-based models have demonstrated strong representation learning capabilities for CMR analysis; however, they typically learn unified latent embeddings that entangle view-specific anatomical variations with disease-related features. Such entanglement biases classifiers toward structural attributes rather than view-invariant pathological patterns. This issue is exacerbated in low-data regimes, particularly for underrepresented cardiac conditions, where limited samples increase the susceptibility to shortcut learning and view-dependent decision boundaries. To address this, we propose a Motion-Guided View--Disease Disentanglement framework (\textbf{MoViD}) built upon a ViT-MAE backbone. The model explicitly factorizes latent representations into view-specific and disease-discriminative components using dual-branch supervised contrastive objectives \textbf{and a gradient-reversal adversarial constraint that minimizes disease leakage into the view embedding}. Additionally, an annotation-free temporal motion feature, derived from inter-frame difference maps, is introduced to localize the beating heart region and suppress background artifacts. A focal reweighting mechanism is incorporated into the contrastive loss to mitigate class imbalance. We evaluate the framework on a private clinical venous thrombosis dataset and two public benchmarks (M\&Ms, M\&Ms2). Across disease classification and cardiac segmentation tasks, our approach consistently outperforms standard transformer baselines and demonstrates competitive performance against large-scale pretrained foundation models, validating the efficacy of structural disentanglement in medical image analysis.

\end{abstract}
%

\section{Introduction}

Cardiac magnetic resonance (CMR) imaging plays a central role in noninvasive assessment of cardiac structure and function. Multiple standardized long-axis views (two-chamber, three-chamber, four-chamber) provide complementary perspectives on the same underlying pathology and support diagnosis across a range of conditions---from common cardiomyopathies to rare, underrepresented diseases where early identification from routine imaging could guide preventive therapy~\cite{garcia2019amyloidosis_review}. Despite this clinical value, the multi-view nature of cine CMR poses a persistent challenge for learning-based methods: view-specific anatomy dominates the learned representation, while disease appearance differs substantially across views. Models trained on multi-view inputs tend to exploit view-dependent shortcuts rather than stable disease cues, degrading generalization when view availability or distributions shift.

Self-supervised foundation models partially mitigate annotation scarcity; CineMA~\cite{cinema} showed that masked autoencoders pretrained on large-scale CMR repositories yield strong transferable features. However, such repositories consist predominantly of healthy or common-condition subjects, meaning the pretrained representations capture normal cardiac morphology and motion far better than rare pathological patterns. When downstream data is scarce---as in low-prevalence conditions like venous thromboembolism (VTE)---the model preferentially retains easier view-dependent features over subtle disease cues, making diagnosis unreliable. The core problem is that \emph{not all information in the representation is useful for diagnosis}; by explicitly decomposing representations into what-to-use (disease) and what-to-ignore (view), we recast the representation learning problem as one of causal factor disentanglement\footnote{We use ``causal'' in the sense of enforcing conditional independence between disease and view factors, analogous to blocking confounding paths in a causal DAG. The gradient-reversal mechanism approximates the interventional distribution $p(Z_d \mid \text{do}(V))$ by ensuring disease embeddings are invariant to view assignment.}.

Several recent works have advanced CMR diagnosis: Martini et al.~\cite{martini2020dl_amyloidosis} and Duffy et al.~\cite{duffy2022cine_amyloidosis} demonstrated CNN-based amyloidosis detection without multi-view integration; Jacob et al.~\cite{jacob2024cmr_foundation} applied DINO pretraining to 36M CMR images; DRIM~\cite{drim2024} separated shared and modality-specific components but targets distinct modalities rather than multi-view acquisitions. None explicitly address view--disease entanglement.

We introduce a motion-guided causal disentanglement framework for multi-view cine CMR that factorizes the representation into a disease embedding (predictive yet view-invariant) and a view embedding (viewpoint-dependent anatomy), enforced through dual-branch contrastive objectives, cross-view consistency regularization, gradient-reversal adversarial constraints, and annotation-free temporal-difference motion cues.

\textbf{Contributions.} (i)~We propose dual-branch contrastive objectives with adversarial independence constraints that disentangle disease-discriminative and view-specific factors, enabling robust classification with limited labeled data and variable view availability. (ii)~We introduce an annotation-free temporal-difference motion cue for cardiac localization that proves to be the most effective individual component in our ablation study. (iii)~We validate on a private venous thrombosis dataset and two public benchmarks (M\&Ms~\cite{campello2021mms}, M\&Ms2~\cite{martinisla2023mms2}), demonstrating competitive or superior performance relative to CineMA~\cite{cinema} without massive pretraining corpora.

\section{Methodology}

\subsection{Overview}

Let $\mathcal{X} = \{x_i^v\}$ denote cine MRI sequences where $x_i^v \in \mathbb{R}^{H \times W \times T}$ is the temporal stack for patient $i$ from view $v \in \mathcal{V} = \{\text{2ch}, \text{3ch}, \text{4ch}\}$, with disease label $y_i \in \{0,1\}$. Our goal is to learn representations satisfying \textit{disease sensitivity} (discriminating pathological from healthy states) and \textit{view invariance} (disease signal independent of acquisition viewpoint).

\begin{figure}[t]
    \centering
    \includegraphics[width=\linewidth]{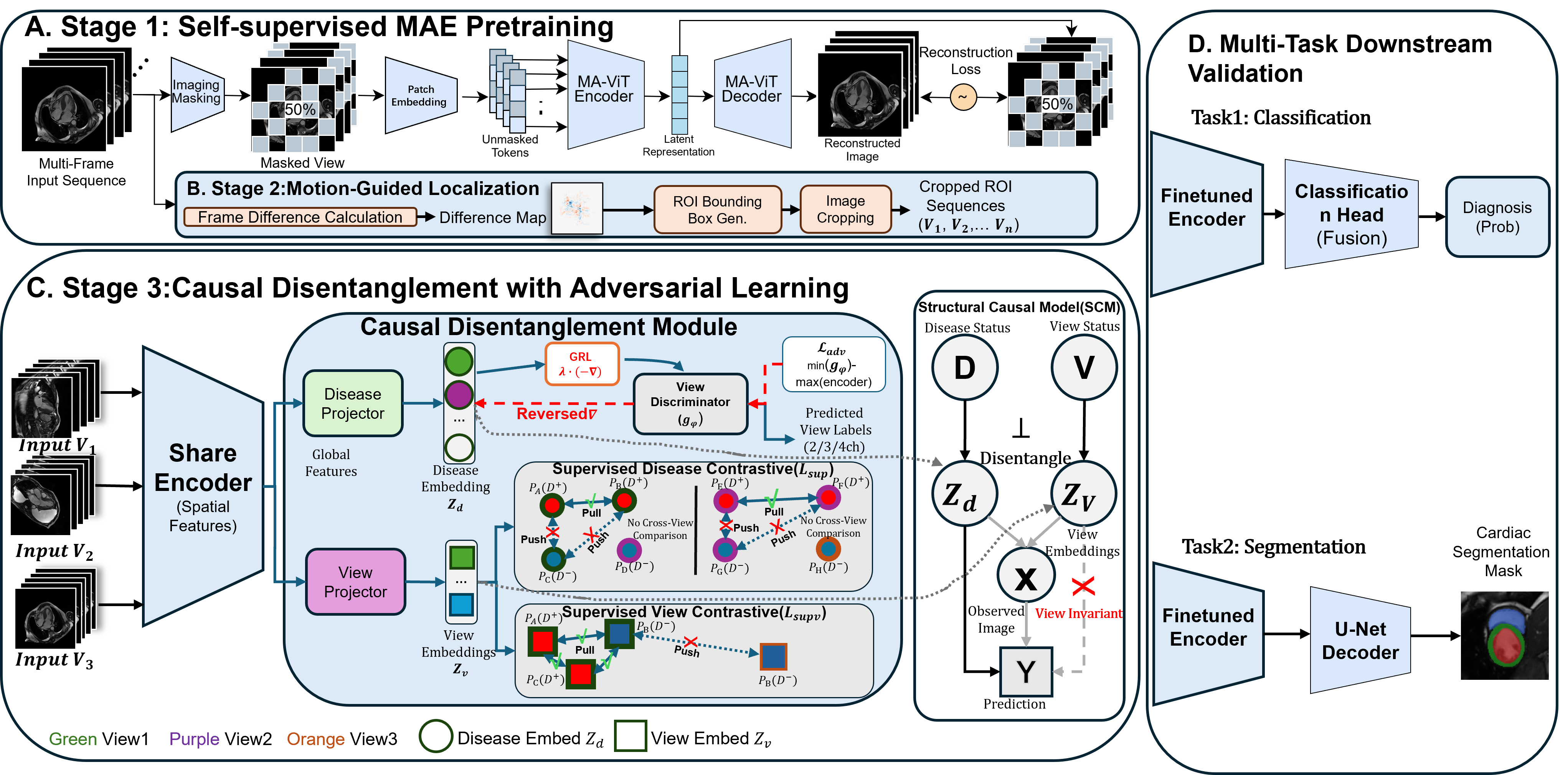}
    \vspace{-5mm}
    \caption{Overall pipeline of MoViD. Stage~A: self-supervised MAE pretraining. Stage~B: temporal-difference cardiac localization. Stage~C: causal disentanglement with dual-branch contrastive learning and adversarial decorrelation. Stage~D: multi-task downstream validation.}
    \label{fig:overview}
\end{figure}

Our framework builds on a convolutional-transformer MAE backbone from CineMA~\cite{cinema}, comprising a 3D CNN stem for spatiotemporal patch tokenization and a shared ViT encoder producing per-view features $\mathbf{f}^v \in \mathbb{R}^D$. The pipeline consists of three stages: (1)~self-supervised MAE pretraining, (2)~motion-guided cardiac localization, and (3)~contrastive fine-tuning with disentangled disease and view embeddings (Figure~\ref{fig:overview}).

\subsection{Self-Supervised Pretraining}

We pretrain the encoder using masked autoencoding~\cite{mae} on unlabeled cine CMR data. 50\% of spatiotemporal patches are randomly masked and the encoder learns to reconstruct the originals through a lightweight transformer decoder using mean squared error. This stage yields rich spatiotemporal representations of cardiac structure and motion without manual annotations. The decoder is discarded after pretraining.

\subsection{Motion-Guided Cardiac Localization}

Cine MRI sequences encode cardiac motion through temporal intensity variations. We exploit this to localize the cardiac region of interest (ROI) without segmentation masks. Given $x \in \mathbb{R}^{H \times W \times T}$, we compute the mean temporal difference map:
\begin{equation}
    \mathbf{D}(h, w) = \frac{1}{T-1} \sum_{t=1}^{T-1} |x_{h,w,t+1} - x_{h,w,t}|
\end{equation}
$\mathbf{D}$ exhibits high values in regions of significant motion (the beating heart) and low values in static structures (chest wall, descending aorta). We threshold $\mathbf{D}$ at the 85th percentile to obtain a binary motion mask, then extract the bounding box of its largest connected component with 15\% adaptive padding as the cardiac ROI. The original sequence is cropped to this bounding box, yielding the encoder input $x' \in \mathbb{R}^{H' \times W' \times T}$ where $H' \!\leq\! H$ and $W' \!\leq\! W$. This annotation-free localization focuses the encoder on diagnostically relevant cardiac structures while eliminating extracardiac confounds such as the chest wall and descending aorta.

\subsection{Disentangled Dual-Branch Representation Learning}

The core contribution lies in the explicit disentanglement of disease and view features through dual projection branches and complementary contrastive objectives.

\subsubsection{Dual-Branch Projection.}
For each per-view feature $\mathbf{f}^v$, two parallel projection heads map to distinct $\ell_2$-normalized embedding spaces: a disease embedding $\mathbf{e}_d^v \in \mathbb{R}^{D_p}$ capturing pathology-relevant features, and a view embedding $\mathbf{e}_v^v \in \mathbb{R}^{D_p}$ encoding view-specific anatomy. Both projectors are two-layer MLPs with batch normalization.

\subsubsection{Disease Contrastive Learning.}
The disease branch uses supervised contrastive learning~\cite{khosla2020supcon} to cluster patients by pathological status \textit{within each view}. Positive pairs share the same disease label within the same view type; negatives have different labels. For view $v$ in a mini-batch:
\begin{equation}
    \mathcal{L}_{\text{disease}}^v = -\frac{1}{|\mathcal{P}_i^v|} \sum_{i} \sum_{p \in \mathcal{P}_i^v} \log \frac{\exp(\mathbf{e}_{d,i}^v \cdot \mathbf{e}_{d,p}^v / \tau_d)}{\sum_{j \neq i} \exp(\mathbf{e}_{d,i}^v \cdot \mathbf{e}_{d,j}^v / \tau_d)}
\end{equation}
where $\mathcal{P}_i^v = \{p : y_p = y_i, p \neq i\}$ and $\tau_d$ is a temperature parameter. The per-view formulation prevents the network from learning shortcuts that exploit view-specific disease patterns. The total disease loss averages $\mathcal{L}_{\text{disease}}^v$ across views.

\textbf{Focal reweighting.} Class imbalance causes underrepresentation of the minority class in negative comparisons. Inspired by~\cite{focal_loss}, we reweight each anchor by positive-pair alignment difficulty:
\begin{equation}
    \mathcal{L}_{\text{disease,focal}}^v = -\frac{1}{|\mathcal{P}_i^v|} \sum_{i} (1 - \bar{s}_i)^\alpha \sum_{p \in \mathcal{P}_i^v} \log \frac{\exp(\mathbf{e}_{d,i}^v \cdot \mathbf{e}_{d,p}^v / \tau_d)}{\sum_{j \neq i} \exp(\mathbf{e}_{d,i}^v \cdot \mathbf{e}_{d,j}^v / \tau_d)}
\end{equation}
where $\bar{s}_i = \frac{1}{|\mathcal{P}_i^v|} \sum_p \mathbf{e}_{d,i}^v \cdot \mathbf{e}_{d,p}^v$ is the mean positive similarity and $\alpha$ is the focusing parameter. Well-clustered anchors ($\bar{s}_i \approx 1$) receive low weight; hard examples from underrepresented classes receive amplified gradients.

\subsubsection{View Contrastive Learning.}
The view branch learns anatomical viewpoint distinctions: positive pairs are samples from the same view type regardless of patient or disease status, while negatives are different views of the same patient:
\begin{equation}
    \mathcal{L}_{\text{view}} = -\sum_{i,v} \log \frac{\sum_{j \neq i} \mathbf{1}[v_j = v] \exp(\mathbf{e}_{v,i}^v \cdot \mathbf{e}_{v,j}^{v_j} / \tau_v)}{\sum_{j \neq i} \exp(\mathbf{e}_{v,i}^v \cdot \mathbf{e}_{v,j}^{v_j} / \tau_v)}
\end{equation}

\subsubsection{Cross-View Consistency Regularization.}
For patients with multiple views, disease embeddings should align (same pathology), while view embeddings should separate (different anatomy):
\begin{equation}
    \mathcal{L}_{\text{consist}} = \sum_{i} \sum_{v_1 \neq v_2} \Big[ \underbrace{(1 - \mathbf{e}_{d,i}^{v_1} \cdot \mathbf{e}_{d,i}^{v_2})}_{\text{disease consistency}} + \underbrace{[\mathbf{e}_{v,i}^{v_1} \cdot \mathbf{e}_{v,i}^{v_2} + \gamma]_+}_{\text{view separation}} \Big]
\end{equation}

\subsubsection{Adversarial View Decorrelation.}
To enforce statistical independence between the two branches, we attach a gradient-reversal layer (GRL)~\cite{ganin2016grl} to the disease branch. A view-prediction head $g_\phi$ attempts to classify the acquisition view from $\mathbf{e}_d^v$:
\begin{equation}
    \mathcal{L}_{\text{adv}} = -\frac{1}{N} \sum_{i} \sum_{v} \mathbf{1}[v_i = v] \log g_\phi(\text{GRL}(\mathbf{e}_{d,i}^v))
\end{equation}
The GRL reverses gradients during backpropagation, encouraging the disease encoder to produce embeddings from which the view cannot be predicted, directly minimizing view-related mutual information in the disease subspace.

\subsubsection{Total Objective.}
The final prediction aggregates view-level logits via late fusion: $\hat{y}_i = \sigma\!\left(\frac{1}{|\mathcal{V}_i|} \sum_{v} \mathbf{w}^\top \mathbf{f}_i^v + b\right)$, trained with cross-entropy $\mathcal{L}_{\text{cls}}$. The complete objective is:
\begin{equation}
    \mathcal{L}_{\text{total}} = \mathcal{L}_{\text{cls}} + \lambda_d \mathcal{L}_{\text{disease}} + \lambda_v \mathcal{L}_{\text{view}} + \lambda_c \mathcal{L}_{\text{consist}} + \lambda_{\text{adv}} \mathcal{L}_{\text{adv}}
\end{equation}

\section{Experiments}

\subsection{Experimental Settings}

\textbf{Datasets.}
We evaluate on three datasets:
(1)~\textit{UVA-Cardiac}: a private cohort of patients with suspected cardiac amyloidosis, imaged with 2ch, 3ch, and 4ch long-axis cine CMR views and binary VTE labels (positive $\approx$9\%).
(2)~\textit{M\&Ms}~\cite{campello2021mms}: multi-vendor SAX cine CMR with segmentation masks (LV, MYO, RV) and five diagnostic categories.
(3)~\textit{M\&Ms2}~\cite{martinisla2023mms2}: SAX and LAX\,4C views; we extract a mid-ventricular and basal SAX slice together with the LAX\,4C view, yielding three input views per subject.
For classification we report ACC, macro F1, AUPRC, AUROC, specificity, and sensitivity; for segmentation, Dice (\%) and 95th-percentile Hausdorff distance (HD95, mm).

\textbf{Baselines and implementation.}
Classification baselines include 3D CNN~\cite{tran2015learning}, ResNet50~\cite{resnet}, ViT-B/16~\cite{dosovitskiy2020vit}, and CineMA~\cite{cinema}. Segmentation baselines include UNet~\cite{ronneberger2015u}, Mamba-UNet~\cite{wang2024mamba}, and CineMA (random-init and fine-tuned). The encoder uses the CineMA ``base'' ViT configuration ($D\!=\!768$, 12 layers, 12 heads). Projection heads are two-layer MLPs (output $D_p\!=\!128$). We use AdamW~\cite{adamw} with cosine decay, batch size 16 on NVIDIA A100 GPUs, and 5-fold stratified cross-validation.

\subsection{Classification Results}

Table~\ref{tab:classification} compares MoViD against baselines across all three datasets. Standard architectures (3DCNN, ResNet50, ViT-B/16) perform near chance on multi-class public benchmarks, confirming that cardiac classification from raw cine sequences remains fundamentally challenging. CineMA's MAE pretraining offers a meaningful advantage, yet its embeddings still entangle view and disease information. MoViD achieves the best results on all datasets: AUROC rises from 0.5382 to 0.7493 on UVA-Cardiac (+39\,pp), from 0.5528 to 0.5963 on M\&Ms, and from 0.6189 to 0.7417 on M\&Ms2, with 0.9956 sensitivity on UVA-Cardiac---critical for VTE screening. Lower absolute accuracy on public benchmarks reflects multi-class classification with severe class imbalance; AUROC, which is threshold-independent, shows consistent improvements across all datasets.

\begin{table}[t!]
\centering
\caption{Classification performance comparison across three datasets. Best results per dataset in \textbf{bold}.}
\label{tab:classification}
\setlength{\abovecaptionskip}{2pt}
\setlength{\belowcaptionskip}{-2pt}
\resizebox{\columnwidth}{!}{%
\scriptsize
\begin{tabular}{@{}llcccccc@{}}
\toprule
\textbf{Dataset} & \textbf{Method} & \textbf{ACC} & \textbf{F1} & \textbf{AUPRC} & \textbf{AUROC} & \textbf{Spec} & \textbf{Sens} \\
\midrule
\multirow{5}{*}{\shortstack[l]{UVA-Cardiac\\(VTE)}}
& 3DCNN~\cite{tran2015learning} & 0.4812 & 0.4173 & 0.5219 & 0.4984 & 0.3004 & 0.8266 \\
& ResNet50~\cite{resnet} & 0.4871 & 0.4289 & 0.0799 & 0.5017 & 0.3018 & 0.8491 \\
& ViT-B/16~\cite{dosovitskiy2020vit} & 0.4983 & 0.4412 & 0.1024 & 0.5198 & 0.3187 & 0.8534 \\
& CineMA~\cite{cinema} & 0.5231 & 0.4675 & 0.5362 & 0.5382 & 0.3621 & 0.8972 \\
& \textbf{MoViD (Ours)} & \textbf{0.6397} & \textbf{0.7033} & \textbf{0.5892} & \textbf{0.7493} & \textbf{0.4812} & \textbf{0.9956} \\
\midrule
\multirow{5}{*}{\shortstack[l]{M\&Ms\\(SAX)}}
& 3DCNN~\cite{tran2015learning} & 0.2433 & 0.1386 & 0.0823 & 0.5322 & 0.0891 & 0.1928 \\
& ResNet50~\cite{resnet} & 0.2401 & 0.1398 & 0.0811 & 0.5381 & 0.0921 & 0.1943 \\
& ViT-B/16~\cite{dosovitskiy2020vit} & 0.2518 & 0.1403 & 0.0841 & 0.5426 & 0.0952 & 0.1987 \\
& CineMA~\cite{cinema} & 0.2692 & 0.1421 & 0.0872 & 0.5528 & 0.0989 & 0.2078 \\
& \textbf{MoViD (Ours)} & \textbf{0.3289} & \textbf{0.1981} & \textbf{0.1089} & \textbf{0.5963} & \textbf{0.1273} & \textbf{0.2561} \\
\midrule
\multirow{5}{*}{\shortstack[l]{M\&Ms2\\(SAX+LAX)}}
& 3DCNN~\cite{tran2015learning} & 0.2596 & 0.1788 & 0.0979 & 0.5668 & 0.1073 & 0.2237 \\
& ResNet50~\cite{resnet} & 0.2655 & 0.1862 & 0.1017 & 0.5736 & 0.1201 & 0.2291 \\
& ViT-B/16~\cite{dosovitskiy2020vit} & 0.2731 & 0.1924 & 0.1063 & 0.5812 & 0.1268 & 0.2348 \\
& CineMA~\cite{cinema} & 0.2982 & 0.2178 & 0.1281 & 0.6189 & 0.1576 & 0.2544 \\
& \textbf{MoViD (Ours)} & \textbf{0.3508} & \textbf{0.2623} & \textbf{0.1972} & \textbf{0.7417} & \textbf{0.2104} & \textbf{0.2851} \\
\bottomrule
\end{tabular}
}
\vspace{-4mm}
\end{table}

\subsection{Ablation Study}

Table~\ref{tab:ablation} reports a progressive ablation on UVA-Cardiac. Adding disease contrastive learning with MAE initialization improves AUROC from 0.5382 to 0.6748; the view branch raises it to 0.6992 by separating viewpoint encoding. Adversarial decorrelation and cross-view consistency yield a further 1.3\,pp gain (0.7126). The largest individual improvement comes from temporal-difference motion cues (AUROC\,=\,0.7369), suggesting that \emph{where you look} matters as much as \emph{how you learn}. The full MoViD model with motion-guided ROI cropping and focal reweighting achieves the best results (AUROC\,=\,0.7493, sensitivity\,=\,0.9956).

\begin{table}[t!]
\centering
\caption{Progressive ablation on UVA-Cardiac (VTE). Each row adds one component to the preceding configuration. Best results in \textbf{bold}.}
\label{tab:ablation}
\setlength{\abovecaptionskip}{2pt}
\setlength{\belowcaptionskip}{-2pt}
\resizebox{\columnwidth}{!}{%
\scriptsize
\begin{tabular}{@{}lcccccc@{}}
\toprule
\textbf{Configuration} & \textbf{ACC} & \textbf{F1} & \textbf{AUPRC} & \textbf{AUROC} & \textbf{Spec} & \textbf{Sens} \\
\midrule
CineMA backbone~\cite{cinema} & 0.5231 & 0.4675 & 0.5362 & 0.5382 & 0.3621 & 0.8972 \\
\quad + Disease Contrastive (no pretrain) & 0.5374 & 0.5802 & 0.5397 & 0.6355 & 0.3061 & 0.8862 \\
\quad + Disease Contrastive (MAE) & 0.5689 & 0.6102 & 0.5413 & 0.6748 & 0.4069 & 0.9262 \\
\quad + View Contrastive & 0.5782 & 0.6278 & 0.5466 & 0.6992 & 0.4128 & 0.9337 \\
\quad + Adversarial + Consistency & 0.5831 & 0.6307 & 0.5486 & 0.7126 & 0.4219 & 0.9481 \\
\quad + Motion Cue (motion only) & 0.6284 & 0.6683 & 0.5532 & 0.7369 & 0.4392 & 0.9762 \\
\quad + Motion-Guided ROI Cropping & 0.6381 & 0.6816 & 0.5653 & 0.7217 & 0.4418 & 0.9833 \\
\quad + Focal Reweighting (\textbf{MoViD}) & \textbf{0.6397} & \textbf{0.7033} & \textbf{0.5892} & \textbf{0.7493} & \textbf{0.4812} & \textbf{0.9956} \\
\bottomrule
\end{tabular}
}
\vspace{-4mm}
\end{table}

\subsection{Segmentation Results}

Table~\ref{tab:segmentation_results} evaluates whether disentangled representations transfer to dense prediction. On M\&Ms, MoViD improves mean Dice by 0.63\,pp over CineMA$^{\text{FineTune}}$ (87.91 vs.\ 87.28\%) and reduces HD95 by 2.05\,mm. On M\&Ms2, MoViD achieves 91.90\% mean Dice and 2.26\,mm HD95, with the largest HD95 improvement on the RV (3.15 vs.\ 7.17/5.22\,mm), suggesting that disentanglement helps attend to irregular boundaries. These gains confirm that the factorized features are not task-specific but yield fundamentally better representations.

\begin{table}[t!]
\centering
\caption{Segmentation performance comparison. Best results in \textbf{bold}.}
\label{tab:segmentation_results}
\setlength{\abovecaptionskip}{2pt}
\setlength{\belowcaptionskip}{-2pt}
\scriptsize
\begin{tabular}{@{}ll|cccc|cccc@{}}
\toprule
\multirow{2}{*}{Dataset} & \multirow{2}{*}{Model} & \multicolumn{4}{c|}{Dice Score (\%) $\uparrow$} & \multicolumn{4}{c}{95\% Hausdorff Distance (mm) $\downarrow$} \\
 &  & RV & MYO & LV & Mean & RV & MYO & LV & Mean \\
\midrule
\multirow{5}{*}{\shortstack[l]{M\&Ms\\(SAX)}}
 & UNet~\cite{ronneberger2015u}   & 86.14 & 82.09 & 89.66 & 85.96 & 6.75 & 6.06 & 6.31 & 6.37 \\
 & Mamba-UNet\cite{wang2024mamba}   & 86.12 & 82.19 & 90.09 & 86.13 & 6.55 & 6.18 & 6.23 & 6.32 \\
 & CineMA$^{\text{RandInit}}$ & 84.65 & 81.12 & 89.23 & 85.00 & 10.26 & 6.25 & 6.80 & 7.77 \\
 & CineMA$^{\text{FineTune}}$ & 87.80 & 83.35 & 90.70 & 87.28 & 5.81 & 5.15 & 5.00 & 5.32 \\
 & \textbf{MoViD (Ours)}               & \textbf{88.19} & \textbf{84.02} & \textbf{91.24} & \textbf{87.91} & \textbf{4.12} & \textbf{3.12} & \textbf{2.41} & \textbf{3.27} \\
\midrule
\multirow{5}{*}{\shortstack[l]{M\&Ms2\\(SAX)}}
 & UNet~\cite{ronneberger2015u}   & 88.26 & 83.56 & 92.25 & 88.02 & 8.63 & 6.49 & 6.37 & 7.16 \\
 & Mamba-UNet\cite{wang2024mamba}   & 88.09 & 82.47 & 93.18 & 87.91 & 8.11 & 5.57 & 6.71 & 6.80 \\
 & CineMA$^{\text{RandInit}}$ & 84.99 & 81.91 & 91.02 & 85.97 & 14.08 & 7.44 & 7.39 & 9.69 \\
 & CineMA$^{\text{FineTune}}$ & 89.72 & 84.80 & 93.12 & 89.21 & 7.17 & 5.31 & 5.27 & 5.91 \\
\midrule
\multirow{4}{*}{\shortstack[l]{M\&Ms2\\(LAX 4C)}}
 & UNet~\cite{ronneberger2015u}   & 90.48 & 86.69 & 94.64 & 90.60 & 5.96 & 4.37 & 4.28 & 4.87 \\
 & CineMA$^{\text{RandInit}}$ & 89.83 & 86.81 & 94.72 & 90.45 & 6.75 & 4.52 & 4.26 & 5.17 \\
 & CineMA$^{\text{FineTune}}$ & 91.09 & 87.64 & 95.59 & 91.44 & 5.22 & 3.61 & 3.39 & 4.07 \\
\midrule
\multirow{1}{*}{\shortstack[l]{M\&Ms2}}
 & \textbf{MoViD (Ours)} & \textbf{91.67} & \textbf{88.04} & \textbf{95.99} & \textbf{91.90} & \textbf{3.15} & \textbf{2.13} & \textbf{1.50} & \textbf{2.26} \\
\bottomrule
\end{tabular}
\vspace{-4mm}
\end{table}

\subsection{Qualitative Analysis}

\begin{figure}[t]
    \centering
    \includegraphics[width=\linewidth]{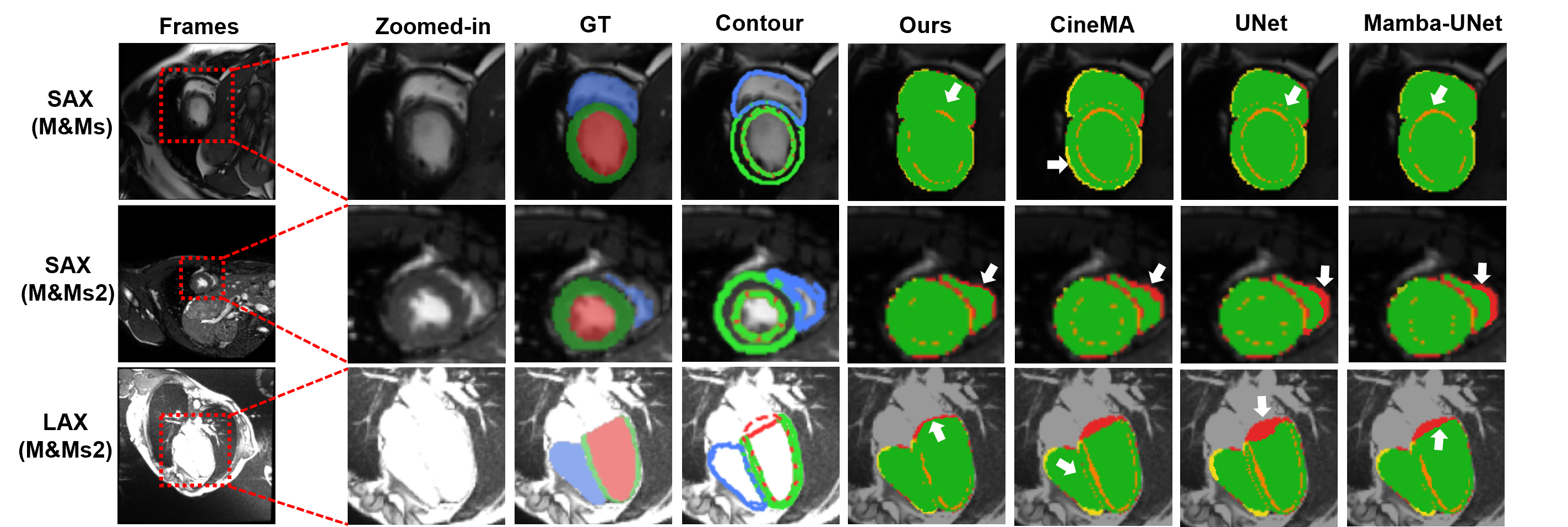}
    \vspace{-5mm}
    \caption{Segmentation comparison on M\&Ms (SAX), M\&Ms2 (SAX), and M\&Ms2 (LAX\,4C). Left to right: frame, zoomed-in region, ground truth, contour overlay, then predictions from MoViD, CineMA, UNet, and Mamba-UNet.}
    \label{fig:seg}
\end{figure}

\begin{figure}[t]
    \centering
    \includegraphics[width=\linewidth]{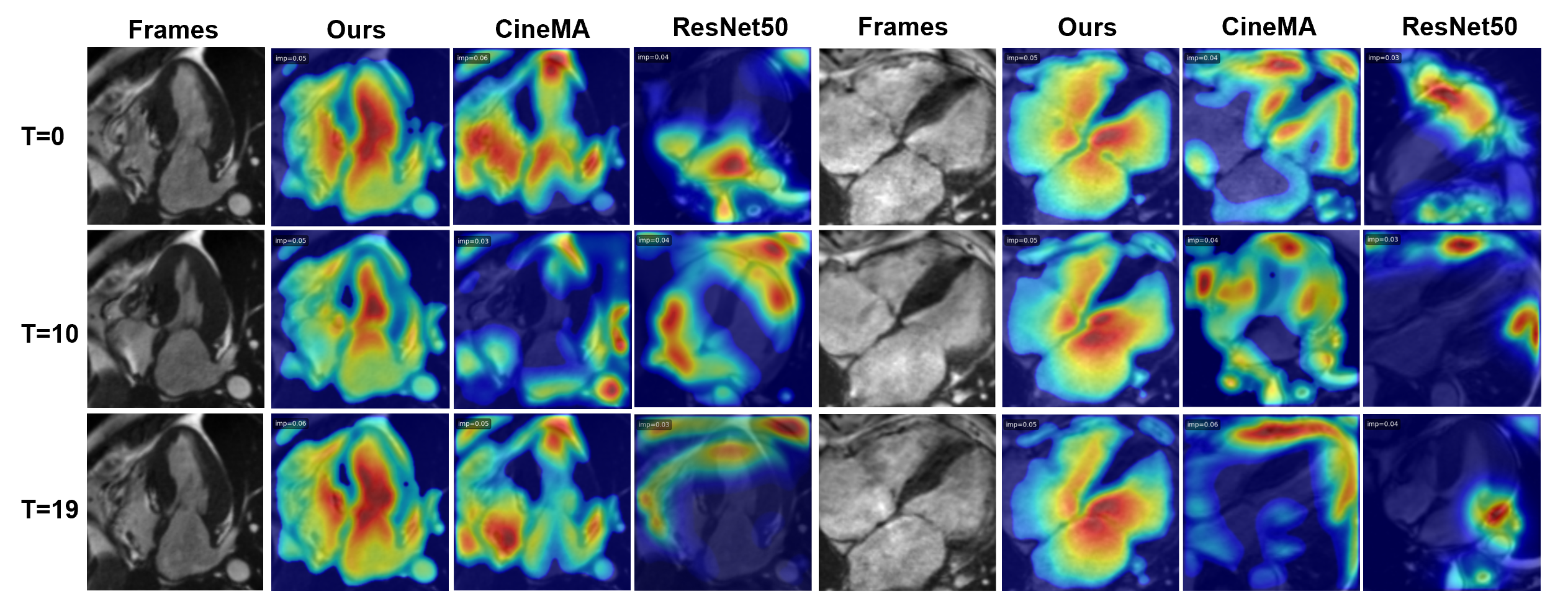}
    \vspace{-5mm}
    \caption{GradCAM attention maps at three temporal frames ($T\!=\!0,10,19$) for two patients. MoViD consistently localizes activation on the cardiac region, whereas CineMA and ResNet50 exhibit diffuse or off-target attention.}
    \label{fig:gradcam}
\end{figure}

Figure~\ref{fig:seg} shows that MoViD delineates the RV free wall and thin myocardial ring more faithfully than CineMA and UNet, which tend to under-segment the myocardium or produce irregular RV boundaries. Figure~\ref{fig:gradcam} visualizes GradCAM activation maps: MoViD consistently concentrates attention within the cardiac region across all time points, whereas CineMA and ResNet50 exhibit diffuse activation on extracardiac structures (descending aorta, chest wall), corroborating the quantitative gains from motion-guided localization.

\section{Conclusion}
We presented MoViD, a motion-guided causal disentanglement framework that factorizes multi-view cine CMR representations into disease-discriminative and view-specific components via dual-branch contrastive objectives, adversarial decorrelation, and annotation-free temporal motion cues. Experiments on a private VTE cohort and two public benchmarks demonstrate consistent improvements in classification and segmentation. The annotation-free motion cue emerged as the most impactful individual component, suggesting that \emph{where you look} may matter as much as \emph{how you learn}. Future work will extend to short-axis views and integrate clinical metadata for multi-modal reasoning.

\begin{credits}
\subsubsection{\ackname} This study was supported by [anonymized for review].

\subsubsection{\discintname}
The authors have no competing interests to declare.
\end{credits}

\bibliographystyle{splncs04}
\bibliography{references}

\end{document}